%% file: main.tex
\documentclass{SPAICE}

\def\authorEmail{angela.cratere@maastrichtuniversity.nl}
\def\AuthorShort{A. Cratere et al.}

\author[1]{Angela Cratere\thanks{Corresponding author. E-Mail: \authorEmail}}
\author[2,3]{Luca Ghilardi}
\author[3, 4]{Vishnu Reddy}
\author[5]{Francesco Dell'Olio}
\author[1]{Charalampos S. Kouzinopoulos}
\author[2,3]{Roberto Furfaro}
\affil[1]{Department of Advanced Computing Sciences, Maastricht University, Maastricht, The Netherlands}
\affil[2]{Systems and Industrial Engineering, University of Arizona, Tucson, USA}
\affil[3]{Space4 Center, University of Arizona, Tucson, USA}
\affil[4]{Lunar and Planetary Laboratory, University of Arizona, Tucson, USA}
\affil[5]{Department of Electrical and Information Engineering, Polytechnic University of Bari, Bari, Italy}

\title{Improving Faint Object Detection for Space Situational Awareness with Variational Autoencoders}

\usepackage[acronym]{glossaries}

\makeglossaries
\glsdisablehyper
\input{acronyms}

\usepackage{makecell}

\begin{document}

\maketitle

\begin{abstract}
We present a deep-learning pipeline for enhancing the detection of faint moving objects in optical space situational awareness (SSA) imagery through automated star removal and background reconstruction. Detecting low-signal-to-noise ratio (SNR) objects remains extremely challenging in optical observations, particularly in the cislunar (X-GEO) environment, where structured sky backgrounds, dense stellar fields, and scattered moonlight significantly degrade the performance of classical detection algorithms. To address this problem, the proposed pipeline combines a lightweight segmentation network (Tiny-U-Net) to generate stellar masks with a partial-convolution variational autoencoder (astro-VAE), designed to learn the statistical distribution of astronomical backgrounds and perform context-aware inpainting of masked regions. The reconstructed background maps can then be used as a preprocessing step to suppress fixed sources and background inhomogeneities prior to detection. As a proof of concept, the approach is integrated with a shift-and-stack scheme and evaluated on real ground-based telescope observations targeting the X-GEO region. Results demonstrate that the method reconstructs star-free backgrounds with high fidelity, while preserving moving targets and significantly enhancing detectability, thereby providing an effective data-driven preprocessing strategy for faint moving-object detection in optical SSA scenarios.
\end{abstract}

\input{Introduction}

\input{Methods}
\input{Results}

\input{Discussion}

\printbibliography
\addcontentsline{toc}{section}{References}

\end{document}

%% file: acronyms.tex
\newacronym{CNN}{CNN}{convolutional neural network}
\newacronym{NN}{NN}{neural network}
\newacronym{DNN}{DNN}{deep neural network}
\newacronym{AI}{AI}{artificial intelligence}
\newacronym{ML}{ML}{machine learning}
\newacronym{DL}{DL}{deep learning}
\newacronym{CAE}{CAE}{convolutional autoencoder}
\newacronym{FC}{FC}{fully connected}
\newacronym{FCN}{FCN}{fully convolutional network}

\newacronym{HW}{HW}{hardware}
\newacronym{FPR}{FPR}{false positive rate}
\newacronym{FNR}{FNR}{false negative rate}
\newacronym{FP}{FP}{false positive}
\newacronym{FN}{FN}{false negative}
\newacronym{BO}{BO}{Bayesian optimization}
\newacronym{TP}{TP}{true positive}
\newacronym{TN}{TN}{true negative}
\newacronym{VAI}{VAI}{Vitis AI}
\newacronym{FP32}{FP32}{32-bit floating-point}
\newacronym{INT8}{INT8}{8-bit integer}
\newacronym{QAT}{QAT}{quantization-aware training}
\newacronym{FLOPs}{FLOPs}{floating-point operations per second}
\newacronym{PR}{PR}{pruning ratio}
\newacronym{IP core}{IP core}{intellectual property core}
\newacronym{MAC}{MAC}{multiply-accumulate}
\newacronym{PTQ}{PTQ}{post-training quantization}
\newacronym{IR}{IR}{intermediate representation}
\newacronym{FPS}{FPS}{frames per second}
\newacronym{mIoU}{mIoU}{mean intersection over union}
\newacronym{IoU}{IoU}{intersection over union}
\newacronym{API}{API}{application programming interface}
\newacronym{HIL}{HIL}{hardware-in-the-loop}

\newacronym{SWaP}{SWaP}{size, weight, and power}
\newacronym{SWaP-C}{SWaP-C}{size, weight, power, and cost}

\newacronym{EC}{EC}{edge computing}
\newacronym{OBC}{OBC}{onboard computer}
\newacronym{FDIR}{FDIR}{fault detection, isolation, and recovery}
\newacronym{CDH}{C\&DH}{command and data handling}
\newacronym{TTC}{TT\&C}{telemetry, tracking, and command}
\newacronym{ACDS}{ACDS}{attitude determination and control system}
\newacronym{TC}{TC}{telecommand}
\newacronym{TM}{TM}{telemetry}
\newacronym{SAVOIR}{SAVOIR}{Space Avionics Open Interface Architecture}

\newacronym{GT}{GT}{ground truth}
\newacronym{SOTA}{SOTA}{state-of-the-art}

\newacronym{COTS}{COTS}{commercial off-the-shelf}
\newacronym{CPU}{CPU}{central processing unit} 
\newacronym{PE}{PE}{processing engine}
\newacronym{DSP}{DSP}{digital signal processor}
\newacronym{muC}{\textmu C}{microcontroller}
\newacronym{FPGA}{FPGA}{field-programmable gate array}
\newacronym{SoC}{SoC}{system-on-a-chip}
\newacronym{GPU}{GPU}{graphics processing unit}
\newacronym{VPU}{VPU}{vision processing unit}
\newacronym{TPU}{TPU}{tensor processing unit}
\newacronym{DPU}{DPU}{deep learning processor unit}
\newacronym{NoC}{NoC}{network-on-chip}
\newacronym{CGRA}{CGRA}{coarse-grained reconfigurable architecture}
\newacronym{PIM}{PIM}{processing-in-memory}
\newacronym{ASIC}{ASIC}{application specific integrated circuit}
\newacronym{SM}{SM}{streaming multiprocessor}
\newacronym{FU}{FU}{functional unit}
\newacronym{CLB}{CLB}{configurable logic block}
\newacronym{LUT}{LUT}{lookup table}

\newacronym{AE}{AE}{autoencoder}
\newacronym{VAE}{VAE}{variational autoencoder}

\newacronym{TOA}{TOA}{top-of-atmosphere}

\newacronym{PSNR}{PSNR}{peak signal-to-noise ratio}
\newacronym{SSIM}{SSIM}{structural similarity index}
\newacronym{RMSE}{RMSE}{root mean squared error}
\newacronym{MSE}{MSE}{mean squared error}
\newacronym{MAE}{MAE}{mean absolute error}

\newacronym{CHIME}{CHIME}{Copernicus Hyperspectral Imaging Mission for the Environment}

\newacronym{Conv2D}{Conv2D}{2D convolution}
\newacronym{Conv2DTranspose}{Conv2DTranspose}{2D transposed convolution}

\newacronym{OA}{OA}{overall accuracy}
\newacronym{macro-F1}{macro-F1}{macro-averaged F1 score}
\newacronym{macro-R}{macro-R}{macro-averaged recall}

\newacronym{macro-IoU}{macro-IoU}{macro-averaged intersection over union}

\newacronym{SSA}{SSA}{space situational awareness}
\newacronym{X-GEO}{X-GEO}{beyond geostationary Earth orbit}
\newacronym{SNR}{SNR}{signal-to-noise ratio}
\newacronym{PC-VAE}{PC-VAE}{partial convolution variational autoencoder}
\newacronym{PCNN}{PCNN}{partial convolution neural network}
\newacronym{UoA}{UoA}{University of Arizona}
\newacronym{PC}{PC}{partial convolution}
\newacronym{ELBO}{ELBO}{evidence lower bound}

\newacronym{KL}{KL}{Kullback–Leibler}
\newacronym{TV}{TV}{total variation}
\newacronym{AFD}{AFD}{active fire detection}

\newacronym{pp}{pp}{percentage points}
\newacronym{NCS2}{NCS2}{Neural Compute Stick 2}
\newacronym{PSF}{PSF}{point spread function}
\newacronym{TBD}{TBD}{track-before-detect}
\newacronym{SandS}{S\&S}{shift-and-stack}
\newacronym{CMB}{CMB}{cosmic microwave background}
\newacronym{NAS}{NAS}{neural architecture search}

\newacronym{NASA}{NASA}{National Aeronautics and Space Administration}

%% file: Introduction.tex
\section{Introduction}\label{sec:introduction}
As the number of active satellites and orbital debris continues to grow \cite{ESA_SpaceEnvironmentReport2025}, \gls{SSA} has become essential for safe and sustainable space operations \cite{Janisch2024IAC_LUCID, Bennett2025OrbitalDebris, LealFilho2025SpaceDebris}. The increasing focus on \gls{X-GEO} missions \cite{Willis2023CislunarCompetition_totheMoon}, partly driven by \gls{NASA}’s Artemis program, has further intensified the need for reliable detection of faint objects under challenging observational conditions. In these scenarios, \gls{SSA} relies heavily on ground-based optical telescopes to detect extremely low-\gls{SNR} targets embedded in complex astronomical backgrounds, affected by instrumental noise, sky gradients, stray light, and dense stellar fields \cite{Popowicz_2015_bkg_estimation}. Accurate background estimation and star removal are therefore critical preprocessing steps in optical \gls{SSA} pipelines. Stellar sources and spatially varying background structures can complicate the detection of faint moving-object signals, and preprocessing must suppress these components without attenuating or removing the target. This is particularly important for integration-based detection methods, where residual stellar contamination and background inhomogeneities can accumulate and generate spurious responses. For instance, detection algorithms such as streak detection \cite{Nir2018StreakDetection_Hough_transform} and \gls{TBD} methods \cite{Nguyen2024FaintObjects_FaXT} are highly sensitive to residual background fluctuations and stellar contamination, which may lead to false positives. Conventional astronomical techniques for background estimation
\cite{Blanton2011SDSSBackground, Popowicz_2015_bkg_estimation,
Liu2023BackgroundModeling, Pandey2025BackgroundEstimation},
source extraction \cite{Stetson_1987_DAOPHOT, refId0_SExtractor},
and source modelling and removal \cite{Liaudat2023PSFmodelling}
remain widely used. However, their performance can depend on manually defined processing choices and parameter settings that may need to be calibrated to the characteristics of each image. This dependence can complicate automated processing across large and heterogeneous \gls{SSA} datasets, where systematic image-by-image inspection and retuning can become impractical at scale. \Gls{DL}-based approaches provide a complementary data-driven strategy by learning the relevant relationships directly from representative observations, reducing the need for image-specific parameter adjustment  and facilitating automated inference. Recent studies have applied \gls{DL} to several astronomical image-processing tasks, including average background prediction \cite{CabayolGarcia2020PAUSurvey_BKGNET}, stray-light suppression \cite{Chen2024StrayLight_PD-LKA}, and image denoising \cite{Vojtekova2021_UNet_denoising, nicolaas2025_BGRem_denoising, Liu2025_nature_Self2Self_denoising}. Notably, the approach in \cite{Madarasz_2025_PConvUNet}, which inspired our work, used a \gls{PCNN} for compact source removal in Herschel far-infrared imagery. However, their method relies on a deterministic architecture and was evaluated on simulated sources. Probabilistic generative models for background reconstruction in real optical \gls{SSA} imagery have remained largely unexplored, particularly when integrated within operational detection pipelines. To address this gap, we propose a \gls{DL} framework for robust background estimation and star removal to enhance faint object detection in \gls{SSA} scenarios. Our objective is to assess whether learned reconstruction can provide an effective preprocessing stage for suppressing structured backgrounds and stellar contamination prior to object detection. The pipeline combines Tiny-U-Net, a customized lightweight U-Net for stellar mask generation, with astro-VAE, a mask-guided \gls{PC-VAE} for background reconstruction. Our main contributions include: (i) the proposed astro-VAE model, enabling probabilistic modeling of structured sky backgrounds and improved suppression of stellar sources compared with deterministic \gls{PCNN}-based approaches, while reducing model complexity; 
(ii) a complete processing pipeline for optical \gls{SSA} imagery combining Tiny-U-Net, astro-VAE, and a detection stage based on a \gls{SandS} scheme, used to evaluate the effect of the proposed preprocessing on faint target visibility;
(iii) validation on real ground-based telescope observations acquired at the \gls{UoA}.

%% file: Methods.tex
\section{Methods}\label{sec:methods}
\begin{figure*}[t!]
    \centering
    \includegraphics[width=\textwidth]{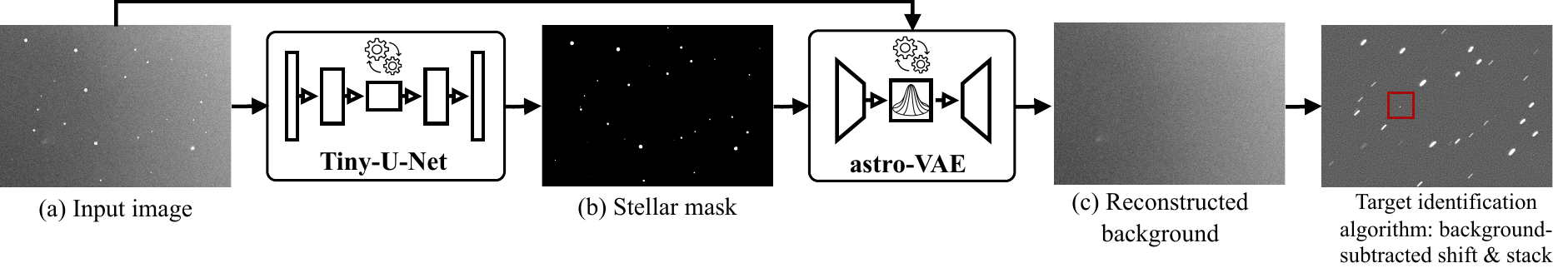}
	\caption{Proposed pipeline: Tiny-U-Net generates a stellar mask (b) from each input image (a); the image and the corresponding mask are fed to the astro-VAE to reconstruct the star-free background (c), which is then used to enhance detection algorithms such as \gls{SandS}.}
	\label{fig:overall_pipeline} 
\end{figure*}

The proposed pipeline integrates (i) a lightweight segmentation network (Tiny-U-Net) for stellar masking and (ii) a variational reconstruction model (astro-VAE) for context-aware background inpainting. Each telescope frame is processed in 256$\times$256 tiles to reconstruct a stellar mask using Tiny-U-Net; then the masked image is provided to astro-VAE to reconstruct a star-free background estimate (Fig.~\ref{fig:overall_pipeline}). The reconstructed background can be used to suppress structured sky variations and stellar clutter before downstream detection algorithms, here integrated with a \gls{SandS} scheme as an initial proof-of-concept implementation \cite{Yanagisawa2001_shift_and_Stack}.

\subsection{Tiny-U-Net} Binary stellar masks are generated using a lightweight version of the U-Net architecture, termed Tiny-U-Net, that we obtained by progressively simplifying a baseline U-Net and comparatively evaluating the two models to quantify the accuracy-complexity trade-off.  The baseline U-Net \cite{U_Net_original_paper_2015} includes four convolutional stages, each composed of two convolutional layers followed by batch normalization and ReLU activations, and a max-pooling operation. The number of filters increases from 64 to 512 (1024-filter bottleneck layer), resulting in 17.27 M parameters. Our Tiny-U-Net employs narrower feature widths (from 16 to 256) and removes batch normalization, with reduced parameter count (1.94 M). This simplification substantially lowers computational complexity and memory requirements. 

\subsection{astro-VAE}
We formulate background reconstruction as a context-aware image inpainting problem \cite{Liu_2018_image_inpainting}, modeling stellar regions as missing data reconstructed from surrounding context. In contrast to \cite{Madarasz_2025_PConvUNet}, which uses a deterministic \gls{PCNN}, we adopt a \gls{PC-VAE}. VAEs extend autoencoders by introducing a probabilistic latent variable model based on variational inference \cite{kingma2013autoencodingvariationalbayes}, enabling the learning of a distribution over latent representations. This choice allows the model to learn the distribution of the structured sky background, capturing its variability and producing smoother and more consistent reconstructions in masked regions. A \gls{PC-VAE} combines variational latent modeling with the spatial selectivity of \gls{PC} layers \cite{Liu_2018_image_inpainting}, which restrict convolutions to valid (unmasked) pixels and propagate the binary mask throughout the network. astro-VAE adopts a U-Net-like architecture with six \gls{PC} blocks in the encoder, each composed of two \gls{PC} layers, with filters increasing from 32 to 256. Strided convolutions reduce the input resolution (256$\times$256$\times$1) to a compact representation (4$\times$4$\times$256), which is flattened and mapped through three fully connected layers to estimate the mean $\mu$ and log-variance $\log\sigma^2$ of a latent vector. The latent variable is sampled using the reparameterization trick, $z = \mu + \sigma \epsilon$, where $\epsilon \sim \mathcal{N}(0,1)$, enabling gradient-based optimization. The sampled latent vector is then passed to a symmetric decoder that reconstructs the image through upsampling and \gls{PC} blocks. The training loss (Eq.~\ref{eq:inpainting_loss}) combines weighted reconstruction terms for both hole ("h") and valid ("v") background regions, perceptual/style, structural similarity, source-consistency, and \gls{KL} regularization: 
\begin{equation}
\label{eq:inpainting_loss}
\begin{aligned}
\mathcal{L}_{\text{train}} =\;&
\lambda_{\text{rec, h}}\,\mathcal{L}_{\text{rec, h}}
+ \lambda_{\text{rec, v}}\,\mathcal{L}_{\text{rec, v}}
+ \lambda_{\text{style}}\,\mathcal{L}_{\text{style}} \\
&+ \lambda_{\text{ssim}}\,\mathcal{L}_{\text{ssim}}
+ \lambda_{\text{source}}\,\mathcal{L}_{\text{source}}
+ \lambda_{\text{KL}}\,D_{\mathrm{KL}} .
\end{aligned}
\end{equation}

The detailed formulations of the first five loss components can be found in prior inpainting works \cite{Liu_2018_image_inpainting, Madarasz_2025_PConvUNet}, whereas the \gls{KL} term arises from our variational formulation. We set the weights to $\lambda_{\text{rec, h}}=6$, $\lambda_{\text{rec, v}}=2$, $\lambda_{\text{style}}=120$, $\lambda_{\text{ssim}}=1$, 
$\lambda_{\text{source}}=1$,
while $\lambda_{\text{KL}}=10^{-3}$ controls the latent space regularization. These values follow common practices in image inpainting \cite{Liu_2018_image_inpainting} and were empirically determined after several experimental trials.

%% file: Results.tex
\section{Results}\label{sec:results}

We evaluated the proposed pipeline on images targeting the \gls{X-GEO} region, acquired by ground-based telescopes operated by the Space4 Center at the \gls{UoA}. The dataset consists of 290 bias-, dark-, and flat-field-corrected optical images (213 used for training and 77 for testing), with varying spatial dimensions (1192$\times$798, 1776$\times$1332, and 2392$\times$1596 pixels). The models were implemented in PyTorch and trained on a workstation equipped with an NVIDIA GeForce RTX 4070 GPU with 12~GB memory.

\subsection{Stellar segmentation}\label{subsec:segmentation_for_stellar_masks}

\begin{figure}[t]
    \centering
    \includegraphics[width=\columnwidth]{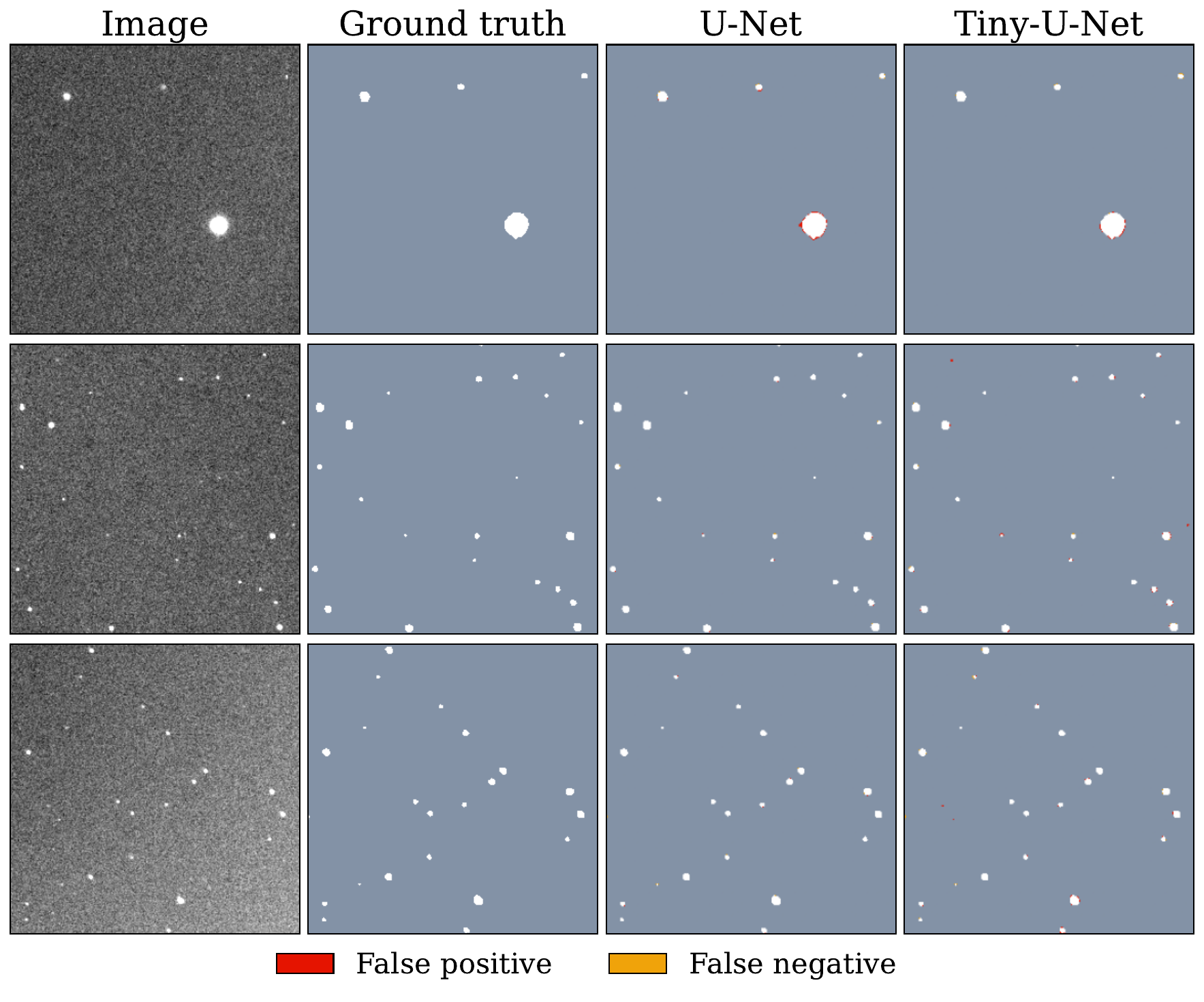}
	\caption{U-Net vs. Tiny-U-Net star segmentation outputs.
	}
	\label{fig:star_segmentation_results}
\end{figure}

To train both Tiny-U-Net and the baseline U-Net, 4,158 monochromatic patches of size 256$\times$256 were extracted (80/20 training-validation split), with ground-truth masks generated using the Astropy Photutils package \cite{larry_bradley_2025_photutils}. Models were trained for 200 epochs using a binary cross-entropy loss. Hyperparameters were optimized via Bayesian search, exploring learning rate ($10^{-5}$–$10^{-2}$), batch size (4, 8, 16), and L2 regularization coefficient ($10^{-8}$–$10^{-2}$), using validation accuracy as the objective. Final evaluation was performed on an independent test set of 1495 patches. Segmentation results are summarized in Table~\ref{tab:unet_metrics_star_segmentation}, while Fig.~\ref{fig:star_segmentation_results} shows some visual examples. While the baseline U-Net achieves slightly higher \gls{IoU} and precision, Tiny-U-Net maintains comparable recall and \gls{OA} with an order-of-magnitude reduction in parameter count (-88.8\%). Given the marginal performance degradation and the significant complexity reduction, Tiny-U-Net was selected for integration into the proposed background estimation pipeline.

\begin{table}
\begin{center}
\begin{tabular}{lcc}
\toprule
\textbf{Model} & \textbf{U-Net} & \textbf{Tiny-U-Net} \\
\midrule
\textbf{IoU$_{\text{star}}$} [\%]& 94.58 & 92.66 \\
\textbf{Precision$_{\text{star}}$} [\%]& 97.80 & 96.05 \\
\textbf{Recall$_{\text{star}}$} [\%]& 96.64 & 96.33 \\
\textbf{mIoU} [\%] & 97.28 & 96.32 \\
\textbf{OA} [\%] & 98.31 & 98.16 \\
\textbf{Params.} [M] & 17.27 & 1.94 \\ 
\bottomrule
\end{tabular}
\caption{Segmentation performance on the test set for U-Net and Tiny-U-Net (threshold = 0.4).}
\label{tab:unet_metrics_star_segmentation}
\end{center}
\end{table}

\subsection{Background reconstruction}
\begin{figure}[t]
    \centering
    \includegraphics[width=.95\columnwidth]{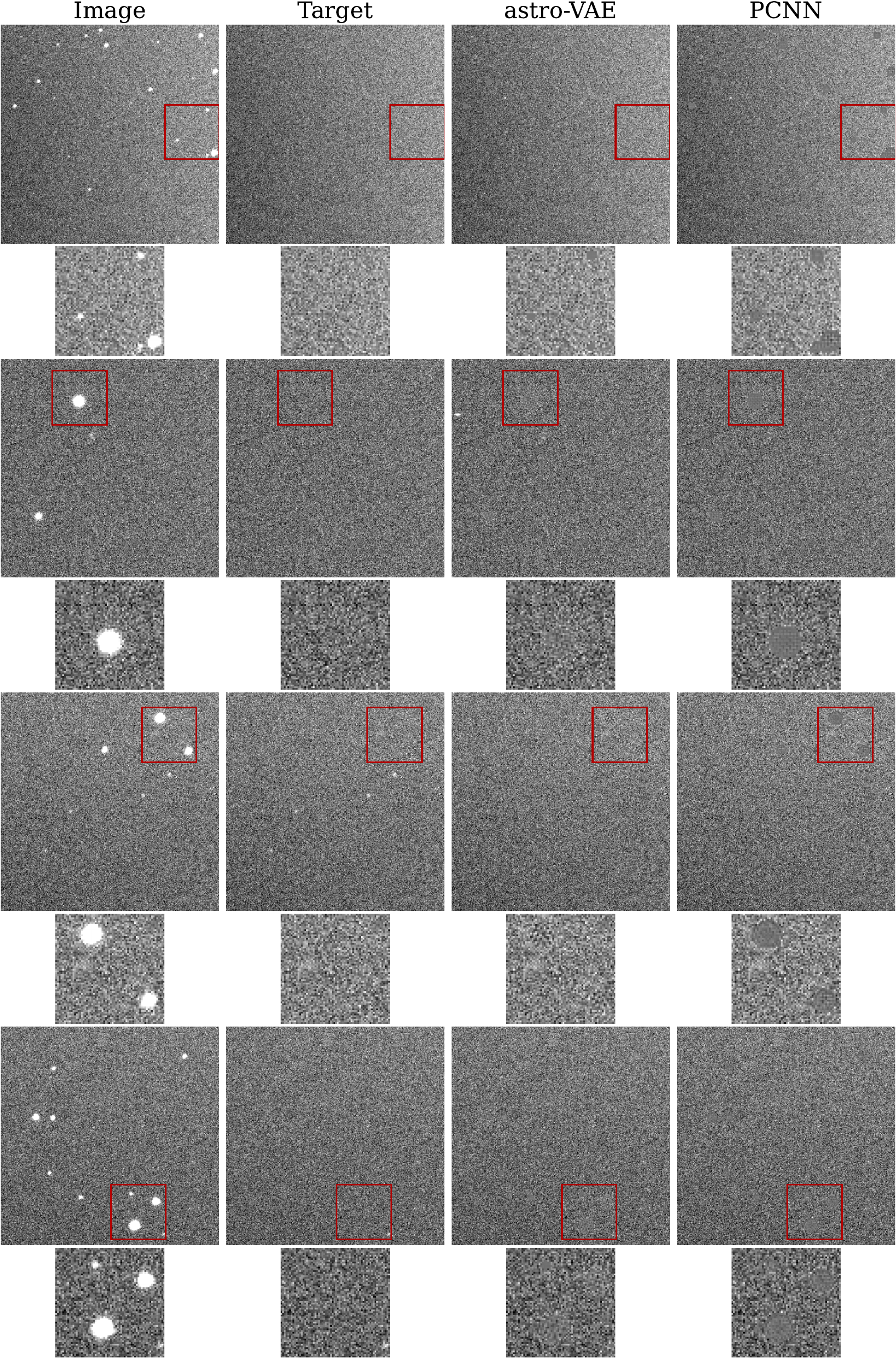}
	\caption{Visual examples of background reconstruction results using astro-VAE. Model predictions are also compared against \gls{PCNN} \cite{Madarasz_2025_PConvUNet}.
	}
	\label{fig:background_reconstruction_results}
\end{figure}

astro-VAE was trained for 300 epochs (learning rate of $10^{-4}$ and batch size 8) in a supervised setting. Because paired observations of the same astronomical scene with and without stellar sources were not available, surrogate star-free reference backgrounds were generated by replacing masked stellar pixels with randomly sampled neighboring background values. These images were used only as supervisory references during training. The network was trained on 10,214 monochromatic patches (80/20 training-validation split) and evaluated on the same independent test set used for segmentation. Reconstruction performance was quantitatively evaluated using \gls{SSIM}, \gls{PSNR}, and \gls{MSE} with respect to the star-free targets. The results are summarized in Table~\ref{tab:astrovae_performance}, while Fig.~\ref{fig:background_reconstruction_results} presents representative qualitative examples. astro-VAE was compared against the deterministic \gls{PCNN} baseline \cite{Madarasz_2025_PConvUNet} trained on the same dataset using an equivalent loss. As reported in Table~\ref{tab:astrovae_performance}, astro-VAE outperforms PCNN across all metrics, while using approximately half the number of total parameters (33.50M vs. 70.13M). Qualitative inspection further confirms that astro-VAE better preserves local background structure while more effectively suppressing stellar residuals, whereas \gls{PCNN} reconstructions exhibit slight texture smoothing in regions previously occupied by bright sources. These results suggest that astro-VAE achieves high reconstruction quality, particularly under structured noise conditions typical of optical \gls{SSA} imagery.

\begin{table}[h]
\begin{center}
\begin{tabular}{lcccc}
\toprule
\textbf{Model} & \textbf{SSIM} & \textbf{PSNR} & \textbf{MSE} & \textbf{Params.} \\
& & [dB] & & [M] \\
\midrule
\textbf{astro-VAE} & \textbf{0.963} & \textbf{35.41} & \textbf{1.31e-3} & \textbf{33.50}\\
PCNN & 0.950 & 31.17 & 2.68e-3 & 70.13\\
\bottomrule
\end{tabular}
\caption{Quantitative background reconstruction results of astro-VAE on the test dataset, compared with the deterministic PCNN baseline. The total parameter count of each model is reported.}
\label{tab:astrovae_performance}
\end{center}
\end{table}

\begin{figure*}[t!]
    \centering
    \includegraphics[width=\textwidth]{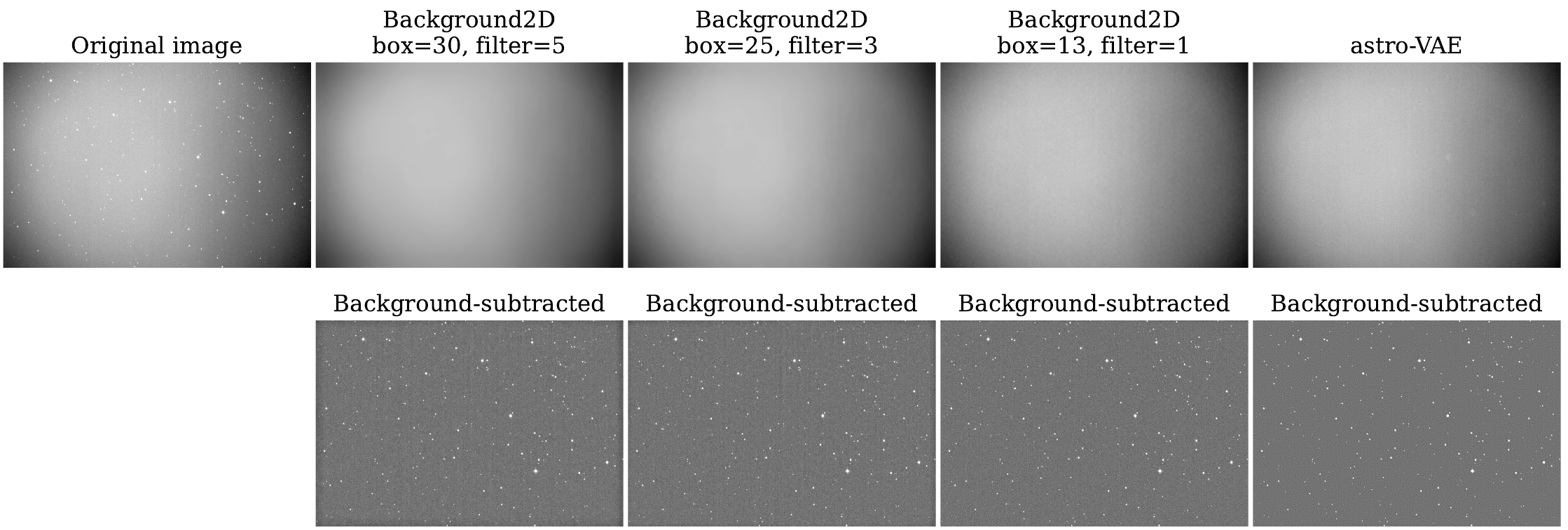}
	\caption{Comparison of conventional and learned background estimation on the same input frame. The conventional estimates are obtained using the \texttt{Background2D} class with different manually selected \texttt{box\_size} and \texttt{filter\_size} configurations. The top row shows the original image, the corresponding \texttt{Background2D} background estimates, and the astro-VAE predicted background. The bottom row shows the associated background-subtracted images.
	}
	\label{fig:astrovae_vs_background2D}
\end{figure*}

Finally, astro-VAE was also compared with a conventional background-estimation method, specifically the \texttt{Background2D} class from Astropy Photutils \cite{larry_bradley_2025_photutils}. This method estimates a spatially varying background from local statistics computed over a predefined mesh and requires the specification of several parameters, including the mesh size, the size of the median filter applied to the low-resolution background map, the source-detection threshold, and the source-mask configuration. Figure~\ref{fig:astrovae_vs_background2D} shows the background estimates obtained for the same input frame while varying the mesh and filter sizes. Larger mesh and filter sizes produce smoother maps dominated by the large-scale background gradient, but leave visible residual structure after subtraction, particularly toward the image borders. Reducing these scales allows the method to follow progressively finer spatial variations and yields a more uniform residual. Thus, the estimated background changes noticeably with the selected configuration, requiring careful calibration of the spatial parameters to the characteristics of the image. In contrast, astro-VAE produces a background estimate using the same learned model without image-specific parameter selection at inference time.

\subsection{Impact on shift-and-stack detection}\label{subsec:integration_tbd}

\begin{table}[]
\begin{center}
\setlength{\tabcolsep}{3pt}
\begin{tabular}{ccc}
\toprule
\textbf{\makecell{Injected SNR \\ (single frame)}} &
\textbf{\makecell{Stacked SNR \\ (standard S\&S)}} &
\textbf{\makecell{Stacked SNR \\ (astro-VAE preproc.)}} \\
\midrule
1.5 & 4.75 & 17.97 \\
2.0 & 5.99 & 25.65 \\
3.0 & 9.11 & 35.40 \\
4.0 & 12.79 & 47.63 \\
5.0 & 14.83 & 64.82 \\
\bottomrule
\end{tabular}
\caption{Stacked SNR comparison with and without astro-VAE preprocessing.}
\label{tab:snr_shiftstack}
\end{center}
\end{table}

\begin{figure*}[t!]
    \centering
    \includegraphics[width=\textwidth]{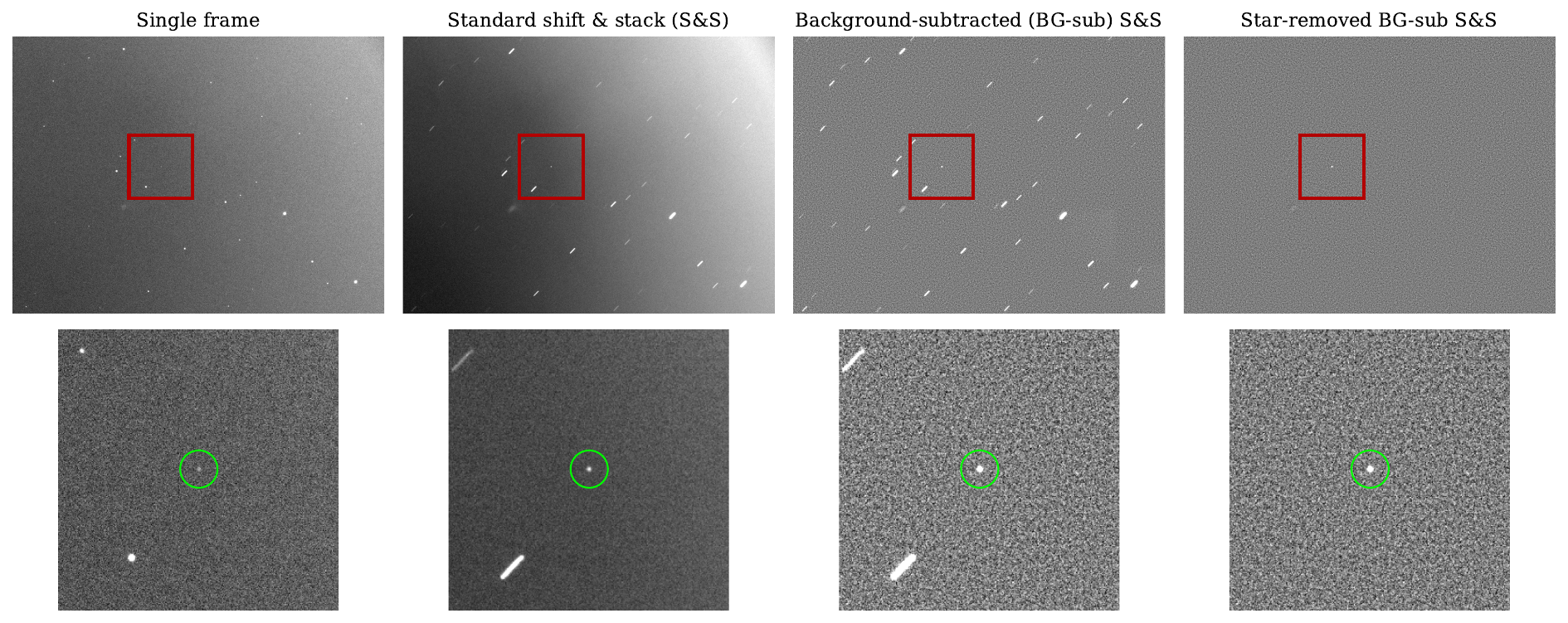}
	\caption{Object detection using \gls{SandS} with and without astro-VAE preprocessing, shown for an injected target with SNR = 3.  From left to right: single frame; standard \gls{SandS}, where background inhomogeneities accumulate, and stars appear as streaks; background-subtracted \gls{SandS} using the astro-VAE reconstructed background, showing improved target contrast; and star-removed background-subtracted \gls{SandS}, demonstrating that the moving target is preserved while fixed stellar sources are suppressed.
	}
	\label{fig:ss_results}
\end{figure*}

We used the reconstructed background maps to enhance target contrast within a simple \gls{SandS} scheme \cite{Yanagisawa2001_shift_and_Stack}, in which the trajectory of the object across successive frames is known. The target signal is coherently integrated along the prescribed track, while background stars appear as streaks due to misalignment. Synthetic point-like targets modeled as 2D Gaussian sources were injected into real telescope frames with \gls{SNR} values between 1.5 and 5, representative of low-\gls{SNR} detection conditions. The target centroid followed a known linear trajectory across 10 frames. We then applied \gls{SandS} in three configurations (Fig.~\ref{fig:ss_results}): (i) stacking the raw injected frames (baseline); (ii) subtracting the astro-VAE background estimate from each frame prior to stacking; and (iii) stacking the inpainted frames, in which stellar sources were removed while preserving the injected target, followed by subtraction of the estimated background map. For faint X-GEO targets, telescopes operate in sidereal tracking mode, pointing toward the predicted target position, where stars remain stationary while the target moves across the detector; therefore the stellar masks in configuration (iii) were generated from the median of multiple frames to avoid removing the target. Results demonstrate that the subtraction of the background map obtained with astro-VAE leads to substantially higher stacked \gls{SNR} values after \gls{SandS}, with improvements of $\approx 4\times$ relative to the baseline \gls{SandS} (Table~\ref{tab:snr_shiftstack}). astro-VAE preprocessing suppresses structured background components and stellar clutter, producing a visibly more uniform residual background after stacking (Fig.~\ref{fig:ss_results}). Importantly, the target is preserved during star removal, enabling reconstruction of star-free backgrounds while retaining moving targets.

%% file: Discussion.tex
\section{Discussion}\label{sec:discussion}

This work addresses a key challenge in optical \gls{SSA} for \gls{X-GEO} surveillance: improving the detectability of extremely faint moving objects embedded in structured astronomical backgrounds. The combined use of Tiny-U-Net and astro-VAE enabled automated stellar masking and context-aware background reconstruction, and the controlled \gls{SandS} experiment showed that this preprocessing can substantially increase the stacked \gls{SNR} of faint injected targets. astro-VAE proved effective in modeling structured sky backgrounds and suppressing stellar sources while preserving moving targets. This capability could be particularly beneficial when the object trajectory is uncertain, as addressed in \gls{TBD} frameworks such as FaXT \cite{Nguyen2024FaintObjects_FaXT}, where multiple velocity hypotheses must be evaluated. In such settings, background structures and fixed stars must be suppressed to avoid coherent integration along incorrect velocity hypotheses, while the unknown moving target must be preserved. Future work will focus on quantifying the impact on false-alarm rate and track purity in \gls{TBD} methods, as well as on model simplification and compression to investigate deployment on embedded systems (e.g., \cite{Cratere_2025}). This would enable low-latency processing directly at the sensor level, near the telescope node, reducing data transmission and accelerating candidate detection and prioritization for follow-up observations in next-generation \gls{SSA} systems.

%% file: library.bib
@article{Madarasz_2025_PConvUNet,
	author = {{Madarász, M.} and {Marton, G.} and {Gezer, I.} and {Lehner, S.} and {Roquette, J.} and {Audard, M.} and {Hernandez, D.} and {Dionatos, O.}},
	title = {A deep neural network approach to compact source removal},
	DOI= "10.1051/0004-6361/202453262",
	url= "https://doi.org/10.1051/0004-6361/202453262",
	journal = {A\&A},
	year = 2025,
	volume = 696,
	pages = "A37",
}

@INPROCEEDINGS{Liu_2018_image_inpainting,
  author    = {G. Liu and F. A. Reda and K. J. Shih and T.-C. Wang and A. Tao and B. Catanzaro},
  title     = {Image inpainting for irregular holes using partial convolutions},
  booktitle = {Proc. Eur. Conf. Comput. Vis. (ECCV)},
  year      = {2018},
  pages     = {89--105},
  publisher = {Springer},
  address   = {Cham, Switzerland},
  doi       = {10.1007/978-3-030-01252-6_6}
}

@ARTICLE{Chen2024StrayLight_PD-LKA,
  author  = {M. Chen and Y. Zhao and W. Yang and others},
  title   = {A model for suppressing stray light in astronomical images based on deep learning},
  journal = {Sci. Rep.},
  volume  = {14},
  pages   = {27521},
  year    = {2024},
  doi     = {10.1038/s41598-024-78472-6},
  url     = {https://doi.org/10.1038/s41598-024-78472-6}
}

@ARTICLE{CabayolGarcia2020PAUSurvey_BKGNET,
  author  = {L. Cabayol-Garcia and M. Eriksen and A. Alarc{\'o}n and A. Amara and J. Carretero and R. Casas and F. J. Castander and E. Fern{\'a}ndez and J. Garc{\'i}a-Bellido and E. Gaztanaga and H. Hoekstra and R. Miquel and C. Neissner and C. Padilla and E. S{\'a}nchez and S. Serrano and I. Sevilla-Noarbe and M. Siudek and P. Tallada and L. Tortorelli},
  title   = {The {PAU} survey: Background light estimation with deep learning techniques},
  journal = {Mon. Not. Roy. Astron. Soc.},
  volume  = {491},
  number  = {4},
  pages   = {5392--5405},
  year    = {2020},
  month   = {02},
  doi     = {10.1093/mnras/stz3274},
  url     = {https://doi.org/10.1093/mnras/stz3274}
}

@ARTICLE{Vojtekova2021_UNet_denoising,
  author  = {A. Vojtekova and M. Lieu and I. Valtchanov and B. Altieri and L. Old and Q. Chen and F. Hroch},
  title   = {Learning to denoise astronomical images with {U\text{-}Nets}},
  journal = {Mon. Not. Roy. Astron. Soc.},
  volume  = {503},
  number  = {3},
  pages   = {3204--3215},
  year    = {2021},
  month   = {05},
  doi     = {10.1093/mnras/staa3567},
  url     = {https://doi.org/10.1093/mnras/staa3567}
}

@misc{nicolaas2025_BGRem_denoising,
      title={{BGRem}: A background noise remover for astronomical images based on a diffusion model}, 
      author={Rodney Nicolaas and Sascha Caron and Fiorenzo Stoppa and Saptashwa Bhattacharyya and Roberto Ruiz de Austri and Paul J. Groot and Andrew J. Levan},
      year={2025},
      eprint={2510.04718},
      archivePrefix={arXiv},
      primaryClass={astro-ph.IM},
      url={https://arxiv.org/abs/2510.04718}, 
}

@ARTICLE{Liu2025_nature_Self2Self_denoising,
  author  = {T. Liu and Y. Quan and Y. Su and others},
  title   = {Astronomical image denoising by self-supervised deep learning and restoration processes},
  journal = {Nat. Astron.},
  volume  = {9},
  pages   = {608--615},
  year    = {2025},
  doi     = {10.1038/s41550-025-02484-z},
  url     = {https://doi.org/10.1038/s41550-025-02484-z}
}

@ARTICLE{Nir2018StreakDetection_Hough_transform,
  author  = {G. Nir and B. Zackay and E. O. Ofek},
  title   = {Optimal and efficient streak detection in astronomical images},
  journal = {Astron. J.},
  volume  = {156},
  number  = {5},
  pages   = {229},
  year    = {2018},
  month   = {10},
  doi     = {10.3847/1538-3881/aaddff},
  url     = {https://doi.org/10.3847/1538-3881/aaddff}
}

@ARTICLE{Nguyen2024FaintObjects_FaXT,
  author  = {T. Nguyen and D. F. Woods and J. Ruprecht and J. Birge},
  title   = {Efficient search and detection of faint moving objects in image data},
  journal = {Astron. J.},
  volume  = {167},
  number  = {3},
  pages   = {113},
  year    = {2024},
  month   = {02},
  doi     = {10.3847/1538-3881/ad20e0},
  url     = {https://doi.org/10.3847/1538-3881/ad20e0}
}

@ARTICLE{Popowicz_2015_bkg_estimation,
  author  = {A. Popowicz and B. Smolka},
  title   = {A method of complex background estimation in astronomical images},
  journal = {Mon. Not. Roy. Astron. Soc.},
  volume  = {452},
  number  = {1},
  pages   = {809--823},
  year    = {2015},
  doi     = {10.1093/mnras/stv1320}
}

@ARTICLE{Pandey2025BackgroundEstimation,
  author  = {P. Pandey and K. Saha},
  title   = {A geometric approach to estimate background in astronomical images},
  journal = {Astrophys. J. Suppl. Ser.},
  volume  = {276},
  number  = {2},
  pages   = {52},
  year    = {2025},
  doi     = {10.3847/1538-4365/ad9906},
  url     = {https://doi.org/10.3847/1538-4365/ad9906}
}

@ARTICLE{Blanton2011SDSSBackground,
  author  = {M. R. Blanton and E. Kazin and D. Muna and B. A. Weaver and A. Price-Whelan},
  title   = {Improved background subtraction for the Sloan Digital Sky Survey images},
  journal = {Astron. J.},
  volume  = {142},
  number  = {1},
  pages   = {31},
  year    = {2011},
  doi     = {10.1088/0004-6256/142/1/31},
  url     = {https://doi.org/10.1088/0004-6256/142/1/31}
}

@ARTICLE{Liu2023BackgroundModeling,
  author  = {Q. Liu and R. Abraham and P. G. Martin and W. P. Bowman and P. van Dokkum and S. R. Janssens and S. Chen and M. A. Keim and D. Lokhorst and I. Pasha},
  title   = {A recipe for unbiased background modeling in deep wide-field astronomical images},
  journal = {Astrophys. J.},
  volume  = {953},
  number  = {1},
  pages   = {7},
  year    = {2023},
  doi     = {10.3847/1538-4357/acdee3},
  url     = {https://doi.org/10.3847/1538-4357/acdee3}
}

@ARTICLE{Yanagisawa2001_shift_and_Stack,
  author  = {T. Yanagisawa and A. Nakajima and T. Kimura and T. Isobe and H. Futami and M. Suzuki},
  title   = {Detection of small {GEO} debris by use of the stacking method},
  journal = {Trans. Jpn. Soc. Aeronaut. Space Sci.},
  volume  = {44},
  number  = {146},
  pages   = {190--199},
  year    = {2001},
  doi     = {10.2322/tjsass.44.190},
  url     = {https://www.jstage.jst.go.jp/article/tjsass/44/146/44_146_190/_article/-char/en}
}

@ARTICLE{Bennett2025OrbitalDebris,
  author  = {M. M. Bennett},
  title   = {Orbital debris requires prevention and mitigation across the satellite life cycle},
  journal = {Commun. Eng.},
  volume  = {4},
  pages   = {95},
  year    = {2025},
  doi     = {10.1038/s44172-025-00430-5},
  url     = {https://doi.org/10.1038/s44172-025-00430-5}
}

@ARTICLE{LealFilho2025SpaceDebris,
  author  = {W. Leal Filho and I. R. Abubakar and J. D. Hunt and M. A. P. Dinis},
  title   = {Managing space debris: Risks, mitigation measures, and sustainability challenges},
  journal = {Sustain. Futures},
  volume  = {10},
  pages   = {100849},
  year    = {2025},
  issn    = {2666-1888},
  doi     = {10.1016/j.sftr.2025.100849},
  url     = {https://www.sciencedirect.com/science/article/pii/S2666188825004149}
}

@ARTICLE{Willis2023CislunarCompetition_totheMoon, 
  author  = {S. Willis},
  title   = {To the {Moon}: Strategic competition in the cislunar region}, 
  journal = {{\AE}ther: A Journal of Strategic Airpower \& Spacepower}, 
  volume  = {2},
  number  = {Spec. Ed.},
  pages   = {17--30},
  year    = {2023},
  note    = {Contribution from Air University Press},
  url     = {https://www.jstor.org/stable/48751535}
}

@ARTICLE{Liaudat2023PSFmodelling,
  author  = {T. I. Liaudat and J.-L. Starck and M. Kilbinger},
  title   = {Point spread function modelling for astronomical telescopes: A review focused on weak gravitational lensing studies},
  journal = {Front. Astron. Space Sci.},
  volume  = {10},
  year    = {2023},
  doi     = {10.3389/fspas.2023.1158213},
  url     = {https://www.frontiersin.org/journals/astronomy-and-space-sciences/articles/10.3389/fspas.2023.1158213/full}
}

@TECHREPORT{ESA_SpaceEnvironmentReport2025,
  author      = {{ESA Space Debris Office}},
  title       = {ESA’s Annual Space Environment Report},
  institution = {European Space Agency},
  year        = {2025},
  url         = {https://www.sdo.esoc.esa.int/environment_report/Space_Environment_Report_latest.pdf},
}

@INPROCEEDINGS{Janisch2024IAC_LUCID,
  author    = {K.-I. Janisch and A. Mastropietro and D. Wischert and F. Rometsch and J. Hoffmann and R. Findlay and S. Aliaga and L. A. C. Pinto and D. Baclet and R. Baptista and D. Betco and R. Bischof and E. Celardo and A. Cratere and others},
  title     = {Detection and tracking of space debris in cislunar environment: A phase 0 mission design},
  booktitle = {Proc. 75th Int. Astronautical Congr. (IAC)},
  year      = {2024},
  pages     = {1188--1204},
  address   = {Milan, Italy},
  month     = {10}
}

@misc{kingma2013autoencodingvariationalbayes,
      title={Auto-Encoding Variational Bayes}, 
      author={Diederik P Kingma and Max Welling},
      year={2013},
      eprint={1312.6114},
      archivePrefix={arXiv},
      primaryClass={stat.ML},
      url={https://arxiv.org/abs/1312.6114}, 
}

@INPROCEEDINGS{U_Net_original_paper_2015,
  author    = {O. Ronneberger and P. Fischer and T. Brox},
  title     = {{U--Net}: Convolutional networks for biomedical image segmentation},
  booktitle = {Proc. Int. Conf. Med. Image Comput. Comput.-Assist. Intervent. (MICCAI)},
  year      = {2015},
  pages     = {234--241},
  publisher = {Springer},
  address   = {Cham, Switzerland},
  doi       = {10.1007/978-3-319-24574-4_28}
}

@software{larry_bradley_2025_photutils,
  author       = {Larry Bradley and
                  Brigitta Sipőcz and
                  Thomas Robitaille and
                  Erik Tollerud and
                  Zé Vinícius and
                  Christoph Deil and
                  Kyle Barbary and
                  Tom J Wilson and
                  Ivo Busko and
                  Axel Donath and
                  Hans Moritz Günther and
                  Mihai Cara and
                  P. L. Lim and
                  Sebastian Meßlinger and
                  Zach Burnett and
                  Simon Conseil and
                  Michael Droettboom and
                  Azalee Bostroem and
                  E. M. Bray and
                  Lars Andersen Bratholm and
                  William Jamieson and
                  Adam Ginsburg and
                  Geert Barentsen and
                  Matt Craig and
                  Sergio Pascual and
                  Shivangee Rathi and
                  Marshall Perrin and
                  Brett M. Morris},
  title        = {astropy/photutils: 2.2.0},
  month        = feb,
  year         = 2025,
  publisher    = {Zenodo},
  version      = {2.2.0},
  doi          = {10.5281/zenodo.14889440},
  url          = {https://doi.org/10.5281/zenodo.14889440},
  swhid        = {swh:1:dir:11159107f27a28985192ed1118b1f2055709d093
                   ;origin=https://doi.org/10.5281/zenodo.596036;visi
                   t=swh:1:snp:ae8c4a55d349d43e53cfe9ce92e678fcfe840f
                   3b;anchor=swh:1:rel:0117f67e8888adcdfc85308287dd9c
                   854b466389;path=astropy-photutils-ffb96c5
                  },
}

@article{Stetson_1987_DAOPHOT,
doi = {10.1086/131977},
url = {https://doi.org/10.1086/131977},
year = {1987},
month = {mar},
publisher = {The Astronomical Society of the Pacific},
volume = {99},
number = {613},
pages = {191},
author = {Stetson, Peter B.},
title = {{DAOPHOT}: A Computer Program For Crowded-field Stellar Photometry},
journal = {Publications of the Astronomical Society of the Pacific},
}

@article{refId0_SExtractor,
	author = {{Bertin, E.} and {Arnouts, S.}},
	title = {{SExtractor}: Software for source extraction},
	DOI= "10.1051/aas:1996164",
	url= "https://doi.org/10.1051/aas:1996164",
	journal = {Astron. Astrophys. Suppl. Ser.},
	year = 1996,
	volume = 117,
	number = 2,
	pages = "393-404",
}

@ARTICLE{Cratere_2025,
  author={Cratere, Angela and Farissi, M. Salim and Carbone, Andrea and Asciolla, Marcello and Rizzi, Maria and Dell’Olio, Francesco and Nascetti, Augusto and Spiller, Dario},
  journal={IEEE Journal on Miniaturization for Air and Space Systems}, 
  title={Efficient {FPGA}-Accelerated Convolutional Neural Networks for Cloud Detection on {CubeSats}}, 
  year={2025},
  volume={6},
  number={3},
  pages={187-197},
  doi={10.1109/JMASS.2025.3533018}}
